\documentclass[runningheads]{llncs}
\usepackage{cite}
\usepackage{amsmath,amssymb,amsfonts,mathtools}
\usepackage{graphicx}
\usepackage{textcomp}
\usepackage{xcolor}
\usepackage{algorithm}
\usepackage[noend]{algpseudocode}
\DeclarePairedDelimiter{\norm}{\lVert}{\rVert}

\begin{document}
%
\title{A fast learning algorithm for One-Class Slab Support Vector Machines}
%
%
\author{Bagesh Kumar \and
Ayush Sinha \and
Sourin Chakrabarti \and
Prof. O.P.Vyas}
%

%
\institute{IIIT Allahabad, India}

\maketitle              
\begin{abstract}
One Class Slab Support Vector Machines (OCSSVM) have turned out to be better in terms of accuracy in certain classes of classification problems than the traditional SVMs and One Class SVMs or even other One class classifiers. This paper proposes fast training method for One Class Slab SVMs using an updated Sequential Minimal Optimization (SMO) which divides the multi variable optimization problem to smaller sub problems of size two that can then be solved analytically. The results indicate that this training method scales better to large sets of training data than other Quadratic Programming (QP) solvers.

\keywords{Support Vector Machine \and One Class Slab Support Vector Machine \and Sequential Minimal Optimization.}
\end{abstract}
\section{Introduction}
With the exponential growth of data in today’s world, the need to analyze, classify and act upon the results in a constrained time is a great challenge. Pattern recognition and classification are arguably the most used operations on real world data. The data present for training our models are not always sufficient and accurate. Closed-set problems deal with classification problems where training data consists of samples spread over all classes which might appear for classification whereas open-set recognition problems deal with training data containing samples from only a fraction of all the classes. Scheirer\cite{b6} discusses the various open-set recognition techniques and compares them. Scheirer\cite{b12} also introduced a novel "1-vs-set machine" which carves out a decision space from the output of traditional SVMs. Best fitting hyperplane\cite{b7} approach is also one of the widely known methods for open-set recognition an object detection. \cite{b29} presents the recent developments in support vector machines. It also gives a comparative review of all the variants of SVMs and the limitations are drawn out.  

Support Vector Machine (SVM)\cite{b10} is a supervised machine learning algorithm that has been proven to be highly effective and efficient in solving classification and regression problems. The algorithm creates a hyperplane separating the data given into classes. It can deal with cases where the data is not perfectly separable by introduction of slack variables in the optimization problem\cite{b8}. Problems with high dimensional feature spaces can be tackled by using suitable Kernels which obey Mercer's theorem\cite{b11}.

However, the traditional SVM needs to label the classes so that they can be classified during the testing and training phases. Hence, they are effective only for closed-set problems and does not perform very well for the open-set problems. Scholkopf\cite{b9} has proposed a variation of the traditional SVM algorithm that would work with samples from only one class. The variation is known as the One Class SVM (OCSVM) which essentially works the same way as traditional SVMs except that it finds the optimal hyperplane considering all the training samples to be of one class and only the origin to be of the other class\cite{b13}. One Class SVMs have been successfully employed in various image processing tasks\cite{b15} and anomaly detection problems\cite{b14}.

Despite the research on the one-class classifiers, they still fail to perform well for the open-set problems. To tackle this, another classifier the One Class Slab SVM (OCSSVM)\cite{b1} was proposed in 2016. It uses OCSVM as the base because it scales well for both linear and non-linear problems. However, the OCSVM keeps the target class on one side and doesn't account for the instances of the negative class to be present on that side of the hyperplane, which can cause high false classifications. Unlike this, the OCSSVM encloses the target class between two hyperplanes (which act as a slab). If the an instance lies inside the slab, it is classifies as a member of the target class otherwise not. The OCSSVM showed better results than other SVMs. It performed comparable to other classifiers, such as support vector data description (SVDD)\cite{b5} and Kernel density estimation methods\cite{b16}. The proposed OCSSVM is seen to be applied in various field such as the anomaly identification in gas-turbines\cite{b17}, finding disciplined medical states and false records in biometry\cite{b18}.

Support Vector Machines are basically trained by solving a complex convex optimization problem with certain constraints which boils down to large quadratic programming problems\cite{b19}. There are several methods to solve QP problems such as Newton based QP solvers\cite{b22}, augmented Lagrangian methods\cite{b20} and primal-dual interior methods\cite{b21}. These methods do not scale well to large datasets while training SVMs due to their high memory requirements and numerous amount of computations. SMO\cite{b3} as proposed by John C. Platt provide a faster method for training SVMs by Dividing the problem into smaller sub problems of size two. SMO basically involved two basic steps: selecting the variables to be optimized according to some heuristic and solving the two variable optimization problem analytically. The SMO algorithm was modified for use with One Class SVMs\cite{b2} and provided great results. In this paper, we modify the SMO algorithm which will enable it to work with OCSSVM and improve its training time. Parallel implementations of the SMO algorithm for training of SVMs in order to provide higher scalability for large datasets have shown great results\cite{b4}.  

There has been several advancements in field of training of support vector machines.\cite{b35} gives a modified SMO approach by computation of the working set by constraining the bounds on results of successive optimization problem computations by an efficient loss function. An SMO algorithm which uses Conjugate gradient descent which results in much lesser number of iterations for convergence \cite{b36}. \cite{b37} proposed faster convergence times by using SSGD with SMO and achieved promising results. \cite{b32} compares the various SMO modifications which revolve around parallel processing to speed up the process. Major modifications have been done around kernel function, division of dataset, hardware architecture etc. One of the implementations as suggested by \cite{b31} involves hardware acceleration and parallelism of FPGA to fasten the processing times. \cite{b4} proposed parallelizing the SMO algorithm by dividing the working set with more processors. Designing SMO on VLSI is another way to quicken the process. \cite{b30} gives an overview of the recent advances in the fields of training of support vector machines. It compares the various techniques used to improve SVM performance such as chunking, data reduction, decomposition etc. They also compare the above optimizations on a number of use cases. \cite{b33} proposed a GPU implementation of SMO for efficient training in case of high dimensional hyper-spectral images. \cite{b34} gives an implementation of PDCO for solving Support Vector Machines. The approach was compared against the various datasets used for benchmarking SVM training algorithms. The results showed that PDCO scaled better with dataset size on a number of occasions. \cite{b38} suggested using LFU caching strategy to to improve SVM training times.

\section{One-Class Slab Support Vector Machines}
The OCSSVM finds two parallel hyperplanes (which acts as a slab) that separate the positive and negative classes. The optimization problem to find the hyperplanes primarily has the normal to the hyperplanes $w$ and the distances of the upper and lower hyperplanes from the origin $\rho_{1}$ and $\rho_{2}$ respectively. The problem is given as:
\begin{alignat}{3}
&\!\min_{\mathbf{w},\rho_{1},\rho_{2},\xi,\bar{\xi}}        &\qquad& \frac{1}{2}\norm{\mathbf{w}}_{2}^2 + \frac{1}{\nu_{1}m}\sum\limits_{i=1}^m \xi_{i} - \rho_{1} + \frac{\varepsilon}{\nu_{2}m}\sum\limits_{i=1}^m \bar{\xi_{i}} + \varepsilon \rho_{2}\\
&\text{subject to} &      & \langle \mathbf{w},\Phi(x_{i})\rangle \geq \rho_{1}-\xi_{i},\xi_{i} \geq 0,\\
&                  &      & \langle \mathbf{w},\Phi(x_{i})\rangle \leq \rho_{2}-\bar{\xi_{i}},\bar{\xi_{i}} \geq 0,\forall i=1,...,m.
\end{alignat}

where $m$ is the size of the feature space, $\xi$ and $\bar{\xi}$ are the slack variables for the upper and lower hyperplanes respectively which handle imperfectly separable cases, $\nu_{1}$ and $\nu_{2}$ control the width of the slab by using the ratio of expected anomalies in the data and $\varepsilon$ determines the impact of the slack variables. $\Phi()$ represents the mapping of the kernel function.

The decision function for each point $x$ is given by :
\begin{alignat}{1}
&\!f(x) = sgn\{(\langle \mathbf{w},\Phi(x) \rangle - \rho_{1})( \rho_{2} - \langle \mathbf{w},\Phi(x) \rangle)\}.
\end{alignat}

We now aim to find the Lagrangian for the above convex optimization problem. The Lagrange multipliers $\alpha_{i}, \beta_{i}, \bar{\alpha_{i}}, \bar{\beta_{i}} \geq 0$ are introduced and the equation is formulated as follows:
\begin{multline}
L(\mathbf{w}, \xi, \bar{\xi}, \boldsymbol{\alpha}, \boldsymbol{\bar{\alpha}}, \boldsymbol{\beta}, \boldsymbol{\bar{\beta}}, \rho_{1}, \rho_{2}) 
= \frac{1}{2}\norm{\mathbf{w}}_{2}^2 + \frac{1}{\nu_{1}m}\sum\limits_{i=1}^m \xi_{i} - \rho_{1} +\\ \frac{\varepsilon}{\nu_{2}m}\sum\limits_{i=1}^m \bar{\xi_{i}} + \varepsilon \rho_{2} - \sum\limits_{i=1}^m \alpha_i (\langle \mathbf{w},\Phi(x_i) \rangle - \rho_1 + \xi_i) -\\
\sum\limits_{i=1}^m \bar{\alpha_i} (\rho_2 + \bar{\xi_i} - \langle \mathbf{w},\Phi(x_i) \rangle) - \sum\limits_{i=1}^m \beta_i\xi_i - \sum\limits_{i=1}^m \bar{\beta_i}\bar{\xi_i}
\end{multline}
Now we find the derivative of the Lagrangian with respect to $w$, $\xi$, $\bar{\xi}$, $\rho_{1}$ and $\rho_{2}$, which will give us:
\begin{alignat}{1}
&\!\frac{dL(\mathbf{w}, \xi, \bar{\xi}, \boldsymbol{\alpha}, \boldsymbol{\bar{\alpha}}, \boldsymbol{\beta}, \boldsymbol{\bar{\beta}}, \rho_{1}, \rho_{2})}{d\mathbf{w}} = \mathbf{w} - \sum\limits_{i=1}^m (\alpha_{i}-\bar{\alpha_{i}})\Phi(x_i) = 0\\
&\frac{dL(\mathbf{w}, \xi, \bar{\xi},\boldsymbol{\alpha}, \boldsymbol{\bar{\alpha}}, \boldsymbol{\beta}, \boldsymbol{\bar{\beta}}, \rho_{1}, \rho_{2})}{d\xi} = \frac{1}{\nu_{1}m} - \alpha_{i} - \beta_{i} = 0 \\
&\frac{dL(\mathbf{w}, \xi, \bar{\xi}, \boldsymbol{\alpha}, \boldsymbol{\bar{\alpha}}, \boldsymbol{\beta}, \boldsymbol{\bar{\beta}}, \rho_{1}, \rho_{2})}{d\bar{\xi}} = \frac{\varepsilon}{\nu_{2}m} - \bar{\alpha_{i}} - \bar{\beta_{i}} = 0\\
&\frac{dL(\mathbf{w}, \xi, \bar{\xi}, \boldsymbol{\alpha}, \boldsymbol{\bar{\alpha}}, \boldsymbol{\beta}, \boldsymbol{\bar{\beta}}, \rho_{1}, \rho_{2})}{d\rho_{1}} = -1 + \sum\limits_{i=1}^m \alpha_{i} = 0\\
&\frac{dL(\mathbf{w}, \xi, \bar{\xi},\boldsymbol{\alpha}, \boldsymbol{\bar{\alpha}}, \boldsymbol{\beta}, \boldsymbol{\bar{\beta}}, \rho_{1}, \rho_{2})}{d\rho_{2}} = -\varepsilon + \sum\limits_{i=1}^m \bar{\alpha_{i}} = 0
\end{alignat}
From the above equations, we could deduce the following results:
\begin{alignat}{1}
&\!\mathbf{w} = \sum\limits_{i=1}^m (\alpha_{i}-\bar{\alpha_{i}})\Phi(x_i) = 0\\
&\beta_{i} = \frac{1}{\nu_{1}m} - \alpha_{i} = 0 \\
&\bar{\beta_{i}} = \frac{\varepsilon}{\nu_{2}m} - \bar{\alpha_{i}} = 0\\
&\sum\limits_{i=1}^m \alpha_{i} = 1\\
&\sum\limits_{i=1}^m \bar{\alpha_{i}} = \varepsilon
\end{alignat}
Substituting equations $(11)$, $(12)$ and $(13)$ into $(5)$, we get the dual problem:
\begin{alignat}{3}
&\!\min_{\boldsymbol{\alpha},\boldsymbol{\bar{\alpha}}}        &\qquad& \frac{1}{2}(\boldsymbol{\alpha}-\boldsymbol{\bar{\alpha}})^{T}K^{T}(\boldsymbol{\alpha}-\boldsymbol{\bar{\alpha}})\\
&\text{subject to} &      & 0 \leq \alpha_{i} \leq \frac{1}{\nu_{1}m},\sum\limits_{j=1}^m \alpha_{j} = 1,\\
&                  &      &0 \leq \bar{\alpha_{i}} \leq \frac{\varepsilon}{\nu_{2}m},\sum\limits_{j=1}^m \bar{\alpha_{j}} = \varepsilon,\forall i=1,...,m.
\end{alignat}
The decision function in terms of $\alpha$ and $\bar{\alpha}$ is:
\begin{multline}
f(x) = sgn\{(\sum\limits_{i=1}^m (\alpha_{i}-\bar{\alpha_{i}})k(x_{i},x) - \rho_{1})
(\rho_{2} -  \sum\limits_{i=1}^m (\alpha_{i}-\bar{\alpha_{i}})k(x_{i},x))\}
\end{multline}
We can recover $\rho_{1}$ and $\rho_{2}$ in terms of the dual variables $\alpha$ and $\bar{\alpha}$ as:
\begin{alignat}{1}
&\!\rho_{1} = \frac{1}{n_{1}}\sum\limits_{i:0<\alpha_{i}<\frac{1}{\nu_{1}m}} \sum\limits_{j=1}^m (\alpha_{j}-\bar{\alpha_{j}})k(x_{i},x_{j})\\
&\rho_{2} = \frac{1}{n_{2}}\sum\limits_{i:0<\bar{\alpha_{i}}<\frac{\varepsilon}{\nu_{2}m}} \sum\limits_{j=1}^m (\alpha_{j}-\bar{\alpha_{j}})k(x_{i},x_{j})
\end{alignat}
where $n_{1}$ and $n_{2}$ are the number of support vectors of the lower and upper hyperplanes respectively.
\section{SMO Algorithm for OCSSVM}
\subsection{Solving reduced optimization problem}
SMO works by breaking a large QP problem into smaller QP Problems, thus decreasing the time required to solve the problem. For SMO, the sub-problems of size 2 can be solved analytically. \\
We have four variables($\alpha_a, \alpha_b, \bar{\alpha_a}, \bar{\alpha_b}$) instead of two as in traditional SMO.
Let $k_{ij}$ denote $k(x_i, x_j)$.
Hence, the QP problem $(16)$ now boils down to:
\begin{multline}
L(\alpha_{a}, \bar{\alpha_{a}},\alpha_{b},\bar{\alpha_{b}}) 
= \frac{1}{2}(\alpha_{a}-\bar{\alpha_{a}})^2k_{aa} + \\\frac{1}{2}(\alpha_{b}-\bar{\alpha_{b}})^2k_{bb}+
(\alpha_{a}-\bar{\alpha_{a}})(\alpha_{b}-\bar{\alpha_{b}})k_{ab}+\\\sum\limits_{i=a,b}((\alpha_i-\bar{\alpha_i})\sum\limits_{j=1,j\neq a,b}^m (\alpha_j-\bar{\alpha_j})k_{ij}) + L\textquotesingle
\end{multline}
where $L\textquotesingle$ is constant with respect to the variables in the above equation. The equality constraints $(14) \ \& \ (15)$, give us:
\begin{alignat}{3}
s^* = \alpha_a^* + \alpha_b^* = \alpha_a + \alpha_b \\
\bar{s}^* = \bar{\alpha_a}^* + \bar{\alpha_b}^* = \bar{\alpha_a} + \bar{\alpha_b}
\end{alignat}
The superscript * used here is to specify the values computed using the old parameter values.
From $(23) \ \& \ (24)$, we have:
\begin{alignat}{3}
\alpha_a = s^* - \alpha_b \\
\bar{\alpha_a} = \bar{s}^* - \bar{\alpha_b}
\end{alignat}
Therefore, $(22)$ now becomes:
\begin{multline}
L(\alpha_b, \bar{\alpha_b}) = \frac{1}{2}(s^* - \bar{s}^* - \alpha_b + \bar{\alpha_b})^2k_{aa} + \\ \frac{1}{2}(\alpha_b - \bar{\alpha_b})^2k_{bb} +
(s^* - \bar{s}^* - \alpha_b + \bar{\alpha_b})(\alpha_b - \bar{\alpha_b})k_{ab} + \\
(\alpha_b - \bar{\alpha_b})\sum_{j=1, j \neq a,b}^m(\alpha_j - \bar{\alpha_j})k_{bj} + \\
(s^* - \bar{s}^* - \alpha_b + \bar{\alpha_b})\sum_{j=1, j \neq a,b}^m(\alpha_j - \bar{\alpha_j})k_{aj} + L\textquotesingle
\end{multline}
To find optimal value for the above equation we need to compute $\alpha_b^*$ and $\bar{\alpha_b}^*$. The partial derivative of equation $(27)$ with respect to $\alpha_b$ and $\bar{\alpha_b}$ gives us:
\begin{multline}
\frac{\partial L(\alpha_b, \bar{\alpha_b})}{\partial\alpha_b} = -1*\frac{\partial L(\alpha_b, \bar{\alpha_b})}{\partial\bar{\alpha_b}} = (\alpha_b+\bar{s}^*-s^*-\bar{\alpha_b})k_{aa}+\\(\alpha_b-\bar{\alpha_b})k_{bb}+(s^*-\bar{s}^*-2\alpha_b+2\bar{\alpha_b})k_{ab}+\\\sum\limits_{j=1,j\neq a,b}(\alpha_j-\bar{\alpha_b})(k_{bj}-k_{aj})
\end{multline}
We know that if $\frac{\partial g(x,y)}{\partial x} = -\frac{\partial g(x,y)}{\partial y}$, then $g(x,y)$ can be expressed as $g(x-y,0)$.
Hence, $L(\alpha_b, \bar{\alpha_b}) = L(\alpha_b-\bar{\alpha_b})$. Let $\gamma = \alpha - \bar{\alpha}$. Hence, equation $(22)$ becomes:
\begin{multline}
L(\gamma_{a},\gamma_{b}) 
= \frac{1}{2}\gamma_a^2k_{aa} + \frac{1}{2}\gamma_b^2k_{bb}+
\gamma_a\gamma_bk_{ab}+\sum\limits_{i=a,b}(\gamma_i\sum\limits_{j=1,j\neq a,b}^m \gamma_jk_{ij}) + L\textquotesingle
\end{multline}
The dual problem $(16)$ and the constraints $(17)$ and $(18)$ in terms of $\gamma$ can be written as:
\begin{alignat}{3}
&\!\min_{\boldsymbol{\gamma}}        &\qquad& \frac{1}{2}\boldsymbol{\gamma}^{T}K^{T}\boldsymbol{\gamma}\\
&\text{subject to} &      & \frac{-\varepsilon}{\nu_{2}m} \leq \gamma_{i} \leq \frac{1}{\nu_{1}m},\forall i=1,...,m,\\
&                  &      &\sum\limits_{j=1}^m \gamma_{j} = 1-\varepsilon.
\end{alignat}
Equation $(29)$ can be rewritten in terms of only one variable, in the same way as discussed above. Let $t^* = \gamma_a^* + \gamma_b^*$. Hence,
\begin{multline}
L(\gamma_b) = \frac{1}{2}(t^* - \gamma_b)^2k_{aa} + \frac{1}{2}\gamma_b^2k_{bb} +
(t^* - \gamma_b )\gamma_bk_{ab} +\\
\gamma_b\sum_{j=1, j \neq a,b}^m\gamma_jk_{bj} +
(t^* - \gamma_b)\sum_{j=1, j \neq a,b}^m\gamma_jk_{aj} + L \textquotesingle
\end{multline}
Partial derivative of $L(\gamma_b)$ with respect to $\gamma_b$ gives us:
\begin{multline}
\frac{\partial L(\gamma_b)}{\partial \gamma_b} = (\gamma_b-t^*)k_{aa}+\gamma_bk_{bb}+(t^*-2\gamma_b)k_{ab}+\sum\limits_{j=1,j\neq a,b}\gamma_j(k_{bj}-k_{aj})
\end{multline}
Putting $\frac{\partial L(\gamma_b)}{\partial \gamma_b} = 0$, we get:
\begin{alignat}{1}
&\!\gamma_b = \gamma_b^* + \eta\sum\limits_{j=1}^m\gamma_j(k_{aj}-k_{bj})\\
&\eta = \frac{1}{k_{aa}+k_{bb}-2k_{ab}}
\end{alignat}
Equation $(35)$ gives us the update rule to update $\gamma_b$ in terms of the old value $\gamma_b^*$. $\gamma_a$ can be updated using the old values of the variables as:
\begin{alignat}{1}
&\!\gamma_a = t^* - \gamma_b
\end{alignat}
To ensure the constraints $(31)$, $(32)$ and $(37)$ for both $\gamma_b$ and $\gamma_a$,we need to define new upper and lower bounds for $\gamma_b$.
\begin{alignat}{1}
&\!L = max(t^*-\frac{1}{\nu_1m},\frac{-\varepsilon}{\nu_2m})\\
&H = min(\frac{1}{\nu_1m},t^*+\frac{\varepsilon}{\nu_2m})
\end{alignat}
\subsection{Variable pair selection heuristic}
SMO uses a heuristic to determine which pair to choose for optimisation. We use  KKT conditions\cite{b23} to check if the current solution is optimal. We determine before each update, if the update to $\gamma_i$ is needed by evaluating KKT conditions. 
\begin{alignat}{1}
\alpha_i(\langle \mathbf{w}, \Phi(x_i)\rangle - \rho_1 + \xi_i) = 0 \\
\beta_i\xi_i = 0 \\
\bar{\alpha_i}(\rho_2 + \bar{\xi_i} - \langle \mathbf{w}, \Phi(x_i)\rangle) = 0 \\
\bar{\beta_i}\bar{\xi_i} = 0 
\end{alignat}
From analysis done in \cite{b1}, we can see that a plane contains a sample if $0 < \alpha_i < \frac{1}{\nu_1m}$ or $0 < \bar{\alpha_i} < \frac{\epsilon}{\nu_2m}$ for lower and upper hyperplanes respectively.
Since a sample can only lie on either upper hyperplane or lower hyperplane, only one of these conditions can be true at a time.
Thus, $\alpha_i > 0$ and $\bar{\alpha_i} > 0$ always occur exclusively.The overall analysis consists of  9 cases. As discussed above, in cases where both $\alpha_i, \bar{\alpha_i} > 0$ indicates that $\rho_2 - \rho_1 \leq 0$ which by the construction of the primal problem should not happen because the slabs overlap and no such slab appears in the feature space. Removing the extra invalid cases, we see that the problem boils down to only 5 cases:
\begin{alignat}{2}
\alpha_i = 0 \ \ and \ \ \bar{\alpha_i} = 0 \ \ and \ \ f(x) > 0\\
0 < \alpha_i < \frac{1}{\nu_1m} \ \ and \ \ \bar{\alpha_i} = 0 \ \ and \ \ f(x) = 0 \\
\alpha_i = \frac{1}{\nu_1m} \ \ and \ \ \bar{\alpha_i} = 0 \ \ and \ \ f(x) < 0 \\
\alpha_i = 0 \ \ and \ \ 0 < \bar{\alpha_i} < \frac{\varepsilon}{\nu_2m} \ \ and \ \ f(x) = 0 \\
\alpha_i = 0 \ \ and \ \ \bar{\alpha_i} = \frac{\varepsilon}{\nu_2m} \ \ and \ \ f(x) < 0 
\end{alignat} 
These equations can be written in terms of $\gamma_i = (\alpha_i - \bar{\alpha_i})$ as follows:
\begin{alignat}{1}
\gamma_i = 0 \ \ \cap \ \ f(x) > 0 \\
\frac{-\varepsilon}{\nu_2m} < \gamma_i < 0 \ \ \cap \ \ f(x) = 0 \\
\gamma_i = \frac{-\varepsilon}{\nu_2m} \ \ \cap \ \ f(x) < 0 \\
0 < \gamma_i < \frac{1}{\nu_1m} \ \ \cap \ \ f(x) = 0 \\
\gamma_i = \frac{1}{\nu_1m} \ \ \cap \ \ f(x) < 0
\end{alignat}

We get the following Lagrange function, considering only the equality constraints from the dual problem in $(30)$, $(31)$ and $(32)$:
\begin{alignat}{3}
\bar{L} = \frac{1}{2}\sum_{i, j = 1}^m\gamma_i\gamma_jk(x_i, x_j) + \lambda(1 - \varepsilon - \sum_{i=1}^m\gamma_i)
\end{alignat}
The gradient of $\bar{L}$ with respect to $\gamma_b$ gives us:
\begin{alignat}{3}
&\!\frac{\partial \bar{L}}{\partial \gamma_b} = \sum_{i=1}^m\gamma_ik_{ib} - \lambda
\end{alignat}
It is known that a function will show maximum variation on change on the variable which has the maximum value of the gradient. According to paper \cite{b24}, the Lagrange multiplier takes the value of the offset $\rho$ in case of One class SVMs. This function basically denotes the distance of the point from the hyper-plane in case of OCSVMs. In our case, the analogue can be defined as the minimum of distances from both hyper-planes. Hence, we define the term $\bar{f(x_b)}$ as:
\begin{alignat}{3}
\bar{f(x_b)} = min(\sum_{i=1}^m\gamma_ik_{ib} - \rho_1, \rho_2 - \sum_{i=1}^m\gamma_ik_{ib})
\end{alignat}
This term would aid us in finding the variable which optimizes the objective function most. Hence, the $\gamma_b$ which gives us $max(|\bar{f(x_b)}|)$. The second variable is chosen as suggested by Scholkopf in \cite{b9}, that is, select $\gamma_a$ which gives us $max(|\bar{f(x_b)}-\bar{f(x_a)}|)$. After selecting the two variables, we use the update rule defined in $(35)$ and $(37)$ and constraints defined in $(38)$ and $(39)$ to get the new values of the variables. We then compute the new values of $\rho_1$ and $\rho_2$. We continue this process until atmost one variable doesn't satisfy the optimality conditions defined in $(49)$, $(50)$, $(51)$, $(52)$ and $(53)$. The algorithm is given in Algorithm 1.
\begin{algorithm}
\caption{Sequential Minimal Optimization for One Class Slab Support Vector Machines}
\textbf{Input:} Training data set
\textbf{Output:} Hyper-plane parametres for classification
\end{algorithm}
\begin{algorithm}
\begin{algorithmic}[1]
\While {More than one variables don't satisfy KKT conditions}
    \State Initialize $\gamma_i$ $\forall$ i in $1,...,m$
    \State Choose $\gamma_b$ which maximizes $|\bar{f(x_b)}|$
    \State Choose $\gamma_a$ which maximizes $|\bar{f(x_b)}-\bar{f(x_a)}|$
    \State Compute new $\gamma_b$ using Equation (35)
    \State Use Equation (38) and (39) to set bounds for $\gamma_b$
    \State Compute new $\gamma_a$ using Equation (37)
    \State Compute $\rho_1$ and $\rho_2$
\EndWhile
\end{algorithmic}
\end{algorithm}
\begin{figure}
\includegraphics[height=200pt]{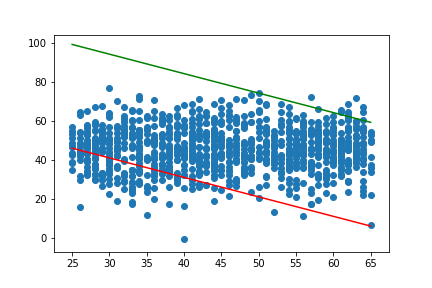}
\caption{The final plot for 1000 data samples of a toy dataset with the data points in blue, the lower and upper hyper-planes in red and green respectively. The horizontal and vertical axes are the x and y axis respectively. The value of constants used were $\nu_1 = 0.5$, $\nu_2 = 0.01$ and $\varepsilon=\frac{2}{3}$.}
\label{fig}
\end{figure}
\begin{figure}
\includegraphics[height=200pt]{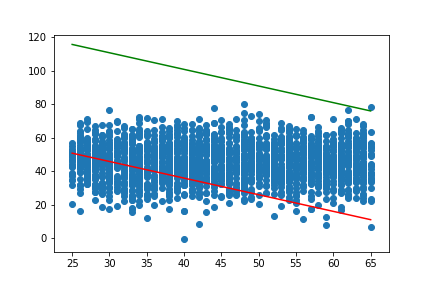}
\caption{The final plot for 2000 data samples of a toy dataset with the data points in blue, the lower and upper hyper-planes in red and green respectively. The horizontal and vertical axes are the x and y axis respectively. The value of constants used were $\nu_1 = 0.2$, $\nu_2 = 0.08$ and $\varepsilon=\frac{1}{2}$.}
\label{fig2}
\end{figure}
\section{Experimentation and Results}
The above algorithm was tested against a toy dataset using a linear kernel $k(x_i,x_j)=x_i \cdot y_i$ and constant values $\nu_1 = 0.5$, $\nu_2 = 0.01$ and $\varepsilon=\frac{2}{3}$. The training times of the algorithm for various sizes of the training dataset was recorded. To find the quality of classification, we recorded the Mathews Correlation Coefficient(MCC)\cite{b28}. The MCC scales well in cases of open set recognition problem datasets. The simulation data in provided in Table \ref{tab1}. Fig. \ref{fig} and Fig. \ref{fig2} shows the final hyperplanes for 1000 and 2000 data samples respectively.

\begin{table}[htbp]
\caption{Training Times and MCC against number of samples}
\begin{center}
\begin{tabular}{|c|c|c|c|c|}
\hline
\textbf{Size}&\textbf{500}&\textbf{1000}&\textbf{2000}&\textbf{5000}\\
\cline{1-5}  
\hline
\textbf{Time(in s)}&0.35&0.67&2.1&5.91\\
\hline
\cline{1-5}
\hline
\textbf{MCC}&0.07&0.13&0.26&0.33\\
\hline
\cline{1-5}
\end{tabular}
\label{tab1}
\end{center}
\end{table}

\section{Conclusion and Future Scope}
This paper introduces an SMO algorithm for training the One Class Slab SVM (OCSSVM). A brief architecture of the OCCSVM was given and the complete derivation inspired by the SMO algrotihm for OCSVMs was proposed. First a reduced optimization problem was derived mathematically followed by devising a suitable working pair selection strategy. Finally, experiments were conducted on the training dataset with variable values of the parameters. The algorithm preformed quite well in terms of accuracy. But the achieved training time was much lower than that of traditional QP solvers which work in complexities which are weakly polynomial\cite{b26}.

Further extensions in this field might include attempts to solve the OCSSVM using other faster methods of training SVMs such as PDCO\cite{b27} or applying parallel SMO\cite{b4}. Improvements in SVM training times can be achieved by implementing hardware level optimizations such as parallelism and caching and by devising a better method for working set selection strategy. Optimizations in loss functions or reducing number of iterations by combining SMO with algorithms like SGD might prove to be of great significance. 

Further, SMO-like algorithms which break down a larger problem into smaller ones should be thought of as the primary solution for any convex optimization problem due to their great advantage of faster training and comparable accuracies over normal quadratic programming solving techniques. Efforts can also be directed towards integrating the algorithm with the popularly available libraries for SVM training.

\end{document}